\documentclass{article} 
\usepackage{fullpage}
\usepackage[outdir=figs/]{epstopdf}
\usepackage{amsmath,amssymb,bm,comment,environ,grffile,latexsym,verbatim,xspace} 

\makeatletter
\@ifpackageloaded{float}{}{
  \usepackage{floatrow}
  \floatsetup[table]{capposition=top}
}
\makeatother

\makeatletter
\@ifpackageloaded{cite}{}{
  \@ifpackageloaded{natbib}{}{\usepackage[sort&compress,numbers]{natbib}}
}
\makeatother

\makeatletter
\@ifpackageloaded{algorithmic}{}{\usepackage[noend]{algpseudocode}}
\@ifpackageloaded{algorithm}{}{\usepackage[ruled]{algorithm}}
\makeatother

\usepackage{graphicx} 
\usepackage{grffile} 

\makeatletter
\@ifclassloaded{beamer}{
  \usepackage[compatibility=false]{caption}  
  }{
  \usepackage{caption}
}
\makeatother

\usepackage[labelformat=simple]{subcaption}
\usepackage{multirow}

\usepackage{tikz}
\usepackage{pgfplots}
\usetikzlibrary{shapes,shapes.arrows,shapes.gates.logic.US,trees,positioning,arrows,automata,decorations.pathmorphing,chains}
\usetikzlibrary{shadows,calc,backgrounds,patterns,fit}
\usetikzlibrary{matrix,arrows}

\usepackage{url}

\makeatletter
\@ifpackageloaded{hyperref}{}{
  \usepackage[colorlinks,linkcolor={red!80!black},citecolor={blue!80!black},urlcolor={blue!80!black}]{hyperref}
}

\@ifclassloaded{beamer}{
  \usepackage{amsfonts}
  }{
  \usepackage{times}
  \usepackage{mathrsfs,dsfont} 
  \usepackage{mathbbol}
  \DeclareSymbolFontAlphabet{\mathbbl}{bbold}
  \usepackage{amsfonts}
  \DeclareSymbolFontAlphabet{\mathbb}{AMSb}

  \usepackage{enumitem}
  \setlist{noitemsep}
  \setlist[1]{noitemsep}
  \setlist[1]{nosep}

  \newlist{compactitem}{itemize}{3}
  \setlist[compactitem]{topsep=0pt,partopsep=0pt,itemsep=0pt,parsep=0pt}
  \setlist[compactitem,1]{label=\textbullet}
  \setlist[compactitem,2]{label=---}
  \setlist[compactitem,3]{label=*}

  \newlist{compactdesc}{description}{3}
  \setlist[compactdesc]{topsep=0pt,partopsep=0pt,itemsep=0pt,parsep=0pt}

  \newlist{compactenum}{enumerate}{3}
  \setlist[compactenum]{topsep=0pt,partopsep=0pt,itemsep=0pt,parsep=0pt}
  \setlist[compactenum,1]{label=\arabic*}
  \setlist[compactenum,2]{label=\alph*}
  \setlist[compactenum,3]{label=\roman*}
}
\makeatother

\usepackage{etex}  
\usepackage{color}
\usepackage{mathtools}
\usepackage{ifthen}
\usepackage{etoolbox}

\usepackage{xr}
\makeatletter
\newcommand*{\addFileDependency}[1]{
        \typeout{(#1)}
        \@addtofilelist{#1}
        \IfFileExists{#1}{}{\typeout{No file #1.}}
}
\makeatother

\input{trmf-def.tex}

\def\trmfsyn{{\sf synthetic}\xspace}
\def\elec{{\sf electricity}\xspace}
\def\traffic{{\sf traffic}\xspace}
\def\wmone{{\sf walmart-1}\xspace}
\def\wmtwo{{\sf walmart-2}\xspace}

\title{High-dimensional Time Series Prediction with Missing Values}

\author{
  Hsiang-Fu Yu\\
  {The University of Texas at Austin} \\
  {rofuyu@cs.utexas.edu}
  \and
  Nikhil Rao\\
{Technicolor Research}\\
{nikhil.rao@technicolor.com}
\and
Inderjit S. Dhillon\\
{The University of Texas at Austin}\\
{inderjit@cs.utexas.edu}
}

\date{}

\newcommand{\T}[1]{\cT_{\text{M}}\rbr{#1\mid \Theta}}

\newcommand{\AR}[1]{\cT_{\text{AR}}\rbr{#1 \mid \cL, \cW, \eta}}
\newcommand{\ARr}[2][]{\cT_{\text{AR}}\rbr{#2 \mid \cL, \bwbar_{#1},\eta}}
\newcommand{\GR}[2][]{\cG\rbr{#2 \mid G^{#1},\eta}}
\newcommand{\GAR}{G^{\text{AR}}}

\newcommand{\Rf}[1]{\cR_f\rbr{#1}}
\newcommand{\Rx}[1]{\cR_x\rbr{#1}}
\newcommand{\Rw}[1]{\cR_w\rbr{#1}}
\newcommand{\Rtheta}[1]{\cR_\theta\rbr{#1}}
\renewcommand{\Prob}[1]{\dP\rbr{#1}}

\newcommand{\best}[1]{{\bf #1}}

\begin{document}

\maketitle

\begin{abstract}
  High-dimensional time series prediction is needed in applications as diverse 
  as demand forecasting and climatology. Often, such applications require methods that are 
both highly scalable, and deal with noisy data in terms of corruptions or 
missing values. Classical time series methods usually fall short of handling 
both these issues. In this paper, we propose to adapt matrix 
matrix completion approaches
that have previously been successfully applied 
to large scale noisy data, but which fail to adequately model 
high-dimensional time series due to temporal dependencies.  We 
present a novel temporal regularized matrix factorization  
(TRMF) framework which supports data-driven temporal dependency learning and 
enables forecasting ability to our new matrix factorization approach. 
TRMF is highly general, and subsumes many existing matrix factorization 
approaches for time series data.  We make interesting connections to graph 
regularized matrix 
factorization methods in the context of learning the dependencies. Experiments 
on both real and synthetic data show that TRMF outperforms several existing 
approaches for common time series tasks. 

\end{abstract}

\section{Introduction}
\label{sec:intro}
Time series data plays a central role in many applications, from demand 
forecasting to speech and video processing to climatology. Such applications involve 
data collected over a large time frame, and also a very large number of 
possibly inter-dependent time series. For example, climatology applications 
involve data collected from possibly thousands of sensors, every 
hour (or less) over several days. Similarly, a store tracking its inventory 
would track thousands of items every day for multiple years. Not 
only is the scale of such problems huge, but they also typically involve 
missing values, due to sensor malfunctions, occlusions or simple human errors. Thus, modern 
time series applications present two challenges to practitioners: scalability 
in the presence of high-dimensional time series and the flexibility to handle missing 
values.  

Classical approaches to handle time-varying data include autoregressive (AR) 
models or dynamic linear models (DLM)~\cite{RK60a,MW13a}. These approaches 
fall short  of handling the aforementioned issues~\cite{OA15a}. Specifically, 
for $T$ time points of $n$ dimensional time series, an AR model of order $L$ 
requires $O(TL^2n^4 + L^3n^6)$ time to  estimate $O(Ln^2)$ parameters, which 
is prohibitive even for moderate values of $n$. Similarly, Kalman Filter based 
DLM approaches need $O(kn^2T + k^3T)$ operations to update parameters, where 
$k$ is the latent dimensionality which may be larger than $n$~\cite{GP09a}.  
As a specific example, the maximum likelihood estimator implementation in the 
widely used R-DLM package~\cite{GP10a}, which relies on a general optimization 
solver, cannot scale beyond $n\ge 32$. (See Appendix~\ref{sec:rdlm} for details).   

Recently, low rank matrix completion or matrix factorization
(MF) methods have found use in large scale collaborative filtering 
~\cite{YK09a,HFY14a}, multi-label learning~\cite{HFY14b}, and many 
other applications where data often scales into the millions. A natural way 
to model high-dimensional time series data is in the form of a matrix, with rows corresponding to time 
series and columns corresponding to time points. In light of this, it is prudent to ask 
%

\begin{quote}
  { \it ``Can we generalize matrix factorization to analyze large scale data with temporal 
  dependencies?'' } 
\end{quote}

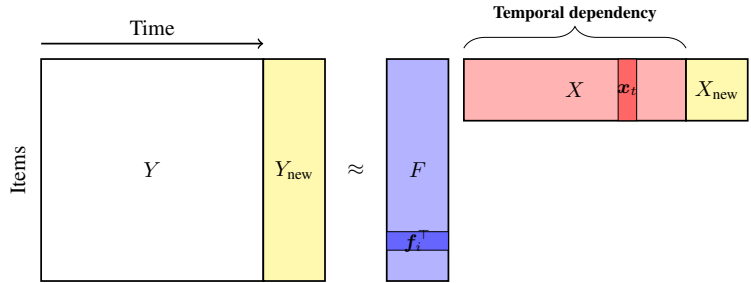
\begin{figure}[!h]
  \begin{center}
    \begin{resize}{0.6\linewidth}
      \begin{tikzpicture}[scale=0.5]
        \def\tsnum{3}
        \def\tslen{3}
        \def\offset{-2}
        \def\gridwidth{2.4}
        \def\asnwidth{0.6}
        \node[rotate=90] at(-0.75+\offset, 0.5*\gridwidth*\tsnum) {Items};
        \draw[thick] (\offset, 0) rectangle (\gridwidth*\tslen+\offset,\gridwidth*\tsnum) 
        node[midway] {$Y$};

        \path[draw,->,thick] (0+\offset, \gridwidth*\tsnum+0.5) -- (\gridwidth*\tslen+\offset, \gridwidth*\tsnum+0.5);
        \node at (0.5*\gridwidth*\tslen+\offset,\gridwidth*\tsnum+1) {Time};

        \node at (\gridwidth*\tslen+1,0.5*\gridwidth*\tsnum)  {$\mathbf{\approx}$};
        \draw[thick,fill=blue!30!white] (\gridwidth*\tslen+2, 0) rectangle 
        (\gridwidth*\tslen+4,\gridwidth*\tsnum) node[midway] {$F$}; 
        \draw[thick,fill=red!30!white] (\gridwidth*\tslen+4.5, \gridwidth*\tsnum-2) rectangle (2*\gridwidth*\tslen+4.5,\gridwidth*\tsnum) node[midway] {$X$}; 

        \draw [decorate,decoration={brace,amplitude=10pt,raise=4pt},yshift=2pt]
        (\gridwidth*\tslen+4.5,\gridwidth*\tsnum) -- (2*\gridwidth*\tslen+4.5,\gridwidth*\tsnum) node [black,midway, yshift=20pt] {\footnotesize \bf{Temporal dependency}};       
        \draw[fill=blue!60!white] (\gridwidth*\tslen+2,1) rectangle (\gridwidth*\tslen+4,1+\asnwidth)  node[midway] {\small{$\bff_i^\top$}};
        \draw[fill=red!60!white] (\gridwidth*\tslen+9.5,\gridwidth*\tsnum-2) rectangle (\gridwidth*\tslen+9.5+\asnwidth,\gridwidth*\tsnum)  node[midway] {\small{$\bx_t$}};

        \draw[thick,fill=yellow!40!white] (\gridwidth*\tslen+\offset, 0) rectangle (\gridwidth*\tslen+\offset+2,\gridwidth*\tsnum) 
        node[midway] {$Y_{\text{new}}$} ;

        \draw[thick,fill=yellow!40!white] (2*\gridwidth*\tslen+4.5, \gridwidth*\tsnum-2) rectangle (2*\gridwidth*\tslen+4.5+2,\gridwidth*\tsnum) node[midway] {$X_{\text{new}}$};
      \end{tikzpicture}
    \end{resize}
  \end{center}
  \caption{Matrix Factorization model for multiple time series. $F$ 
  captures features for each time series in the matrix $Y$, and $X$ 
captures the latent and  time-varying variables. } 
  \label{fig:trmf}
\end{figure}

Let $Y\in\RR^{n\times T}$ be the observed data matrix of $n$ time series of 
length $T$. 
Many classic time series models can be described in the following 
general form~\cite{GP09a,MW13a}:
\begin{align}
  \by_t &= F \bx_t + \eta_t, \label{eq:dlm-obs}\\
  \bx_t & = M_{\Theta}\rbr{\cbr{\bx_{t-l}: l \in \cL}} + \epsilon_t, \label{eq:dlm-state}
\end{align}
where $\by_t\in\RR^n$ is the snapshot of $n$ time series at the $t$-th time point, 
$F\in\RR^{n\times k}$, $\bx_t\in\RR^k$ is the latent embedding for the $t$-th 
time point, and $\eta_t, \epsilon_t$ are Gaussian noise vectors.  $M_\Theta$ in \eqref{eq:dlm-state} is a time series model parameterized by  $\cL$ and $\Theta$: 
$\cL$ is a set containing each lag index $l$ denoting a dependency between 
$t$-th and $(t-l)$-th time points, while $\Theta$ captures the weighting 
information of temporal dependencies (such as the transition matrix in AR 
models).  When we stack all the  
$\bx_t$s into a matrix $X$ and let $\bff_i^\top$ be the $i$-th row of $F$ as 
shown in Figure~\ref{fig:trmf},   
we clearly see that $Y \approx FX$ and we can attempt to learn the factors in
we clearly see that $Y \approx FX$ and we can learn the factors in
~\eqref{eq:dlm-obs} by considering a  
 standard matrix factorization formulation~\eqref{eq:mf}:
\begin{align}
  \min_{F,X} \quad \sum_{(i,t)\in \Omega} \rbr{Y_{it} - \bff_i^\top \bx_t}^2 + 
  \lambda_f \Rf{F} + \lambda_x \Rx{X},
  \label{eq:mf}
\end{align}
where $\Omega$ is the set of the observed entries, and $\Rf{F}, \Rx{X}$ are 
regularizers for $F$ and $X$ respectively.

This connection made, there now arise two major challenges to extend MF approaches to analyze data with temporal dependencies:
%
\begin{itemize}
  \item How to {\bf describe} and {\bf incorporate} the structure
    of temporal dependencies into the MF formulation? 
  \item How to efficiently {\bf forecast} values of future time points? 
\end{itemize}

It is clear that the common choice of the regularizer $\Rx{X} = \norm{X}_F$ is 
no longer appropriate, as it does not take into account the dependencies among the columns of $X$. 
Most existing MF adaptations to handle temporal dependencies~\cite{ZC05a,YZ09a,LX10a,SR10a,MR12a} graph-based  
approaches, where the dependencies are described by a graph and 
incorporated through a Laplacian
regularizer~\cite{AS03a}. However, graph-based regularization fails in 
cases where there are negative correlations between two time 
points. Furthermore, unlike scenarios where explicit graph information is available  
with the data (such as social network or product co-purchasing graph for recommender systems), explicit 
temporal dependency structures are usually unavailable and has to be inferred or approximated. Hence, this approach requires 
practitioners to either perform a separate procedure to estimate the 
dependencies or consider very short-term dependencies with simple fixed 
weights.  Moreover, existing MF approaches, while yielding good 
estimations for missing values in past time points, are poor in terms of forecasting 
future values, which is crucial in time series analysis.  

\subsection{Our Contributions:}
In this paper, we propose a novel temporal 
regularized matrix factorization (TRMF) framework for data with temporal 
dependencies. TRMF generalizes several existing temporal 
matrix factorization approaches~\cite{ZC05a,YZ09a,LX10a,SR10a,MR12a} and 
connects to some  
existing latent time-series models~\cite{RK60a,LL09a,GP09a,LL11a,MW13a}.  
Our approach not only supports data-driven temporal dependency 
learning but also brings the ability to forecast future values to matrix factorization methods. 
Also, unlike classic time-series models, TRMF easily handles high dimensional time series data even in the presence of many missing 
values. 

Next, based on AR models, we design a novel 
temporal regularizer to encourage temporal  
dependencies among the latent embeddings $\cbr{\bx_t}$. We also make interesting 
connections between the proposed regularizer and graph-based approaches~\cite{AS03a}.
This connection not only leads to better understanding about the 
dependency structure incorporated by our framework but also brings the benefit of using 
off-the-shelf efficient solvers such as GRALS~\cite{NR15a} 
directly to solve TRMF.

The rest of this paper is organized as follows. In Section~\ref{sec:related}, 
we review the existing approaches and their limitations on data with temporal 
dependencies. We present the proposed TRMF framework in
Section~\ref{sec:trmf}, and show that the method is highly general and can be 
used for a variety of time series applications. We introduce a novel AR 
temporal regularizer in  
Section~\ref{sec:armf}, and make connections to graph-based regularization 
approaches. We demonstrate the superiority of the proposed  
approach via extensive experimental results in Section~\ref{sec:exp} and 
conclude the paper in Section~\ref{sec:conclusion}.



\section{Data with Temporal Dependency: Existing Approaches and Limitations}
\label{sec:related}

\subsection{Time-Series Models}
\label{sec:ts-relates}

Models such    
as AR and DLM are not suitable for modern multiple high-dimensional time series 
data (i.e., both $n$ and $T$ are large) due to their inherent computational 
inefficiency (see Section~\ref{sec:intro}).  
To reduce the number of parameters and avoid overfitting in AR models, there have been 
studies with various structured  transition matrices such as low 
rank and sparse matrices~\cite{FH13a,WN14a}. The focus of most of these works is on devising 
better statistical consistency guarantees, and the issue of scalability of AR models remains 
a challenge. On the other hand, it is also challenging for many classic 
time-series models to deal with data with missing 
values~\cite{OA15a}.    

DLM unifies many other 
time series models~\cite{GP09a,MW13a}. The original idea goes 
back to the Kalman filter~\cite{RK60a} in the control theory community. In many
situations where the model parameters such as $F$ in \eqref{eq:dlm-obs} and 
$\Theta,\cL$ in \eqref{eq:dlm-state} are either given or designed by 
practitioners, the Kalman filter approach is used to preform forecasting, while the Kalman smoothing approach is used to impute missing entries. In situations where model parameters are unknown, EM algorithms are 
applied to estimate both the model parameters and latent embeddings for 
DLM~\cite{RS82a,ZG96a,LL09a,LL11a,JS12a}. As most EM approaches for DLM 
contain the Kalman filter as a building block, they cannot scale to very high 
dimensional time series data.  Indeed, as shown in Section \ref{sec:exp}, the popular R package for DLM's does not scale beyond data with a few hundred observations and dimensions.


\subsection{Existing Matrix Factorization Approaches for Time Series}
\label{sec:mf-relates}

In standard matrix factorization~\eqref{eq:mf}, the squared Frobenius norm 
$\Rx{X} = \norm{X}_{F} = \sum_{t=1}^T \norm{\bx_t}^2$ is the usual 
regularizer of choice for $X$. Because squared Frobenius norm assumes no dependencies among 
$\cbr{\bx_t}$, standard MF formulation is {\em invariant to column 
permutation} and not applicable to data with temporal dependencies. Hence most 
existing temporal MF approaches~\cite{ZC05a,YZ09a,LX10a,SR10a,MR12a} turn to 
the framework of  graph-based regularization~\cite{AS03a} for temporally 
dependent $\cbr{\bx_t}$, with a graph encoding the temporal dependencies. 

\paragraph{Graph regularization for temporal dependency.}
The framework of graph-based regularization is an approach to describe and 
incorporate general dependencies among variables. 
Let $G$ be a graph over $\cbr{\bx_t}$ and $G_{ts}$ be the weight between 
the $t$-th node and $s$-th node. A popular regularizer to include as part of 
an objective function is the following: 
\begin{equation}
  \Rx{X} = \GR{X} := \frac{1}{2} \sum_{t\sim s} G_{ts}\norm{\bx_t-\bx_s}^2 + 
  \frac{\eta}{2} \sum_{t} \norm{\bx_t}^2,
  \label{eq:graph-reg}
\end{equation}
where $t\sim s$ denotes the edge between $t$-th node and $s$-th node, and the 
second summation term is used to guarantee strong convexity.  

A large $G_{ts}$ will ensure that $\bx_t$ and $\bx_s$ are close to each other 
in the Euclidean sense, when \eqref{eq:graph-reg} is minimized. Note that to 
guarantee the convexity of $\GR{X}$, the weights $G_{ts}$ need to be non-negative.  

\begin{figure}
  \centering
  \begin{resize}{0.48\linewidth}
    \begin{tikzpicture}
      \node [draw,minimum width=2.5em,minimum height=2.5em, circle] at (0,0) (x0) {\scriptsize{$t$}};
      \foreach \t in {-1,-2,-3,-4,+1} {
        \node [draw,minimum width=2.5em,minimum height=2.5em, circle] at (1.5*\t, 0) (x\t)
        {\scriptsize{$t\t$}};
      }
      \node at (1.5*-4.6,0) {$\cdots$};
      \node at (1.5*1.6,0) {$\cdots$};
      \path (x0) edge [draw=blue,very thick,bend left=30] node [above] {\color{blue}$w_1$} (x-1);
      \path (x0) edge [draw=red,very thick,bend right=30] node [above] {\color{red}$w_4$} (x-4);
      \path (x+1) edge [draw=blue,very thick,bend left=30] node [above] {\color{blue}$w_1$} (x0);
      \path (x+1) edge [draw=red,very thick,bend right=30] node [above] {\color{red}$w_4$} (x-3);
      \path (x-1) edge [draw=blue,very thick,bend left=30] node [above] {\color{blue}$w_1$} (x-2);
      \path (x-2) edge [draw=blue,very thick,bend left=30] node [above] {\color{blue}$w_1$} (x-3);
      \path (x-3) edge [draw=blue,very thick,bend left=30] node [above] {\color{blue}$w_1$} (x-4);
    \end{tikzpicture}
  \end{resize}
  \caption{Graph-based regularization for temporal dependencies. }
  \label{fig:time-dep0}
\end{figure}
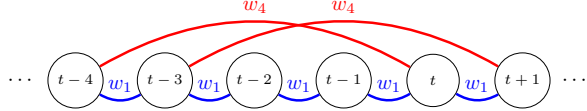

To apply graph-based regularizers to {\em temporal dependencies}, we need to 
specify the (repeating) dependency pattern by a lag set $\cL$ and a weight vector 
$\bw$ such that all the edges $t\sim s$ of distance $l$ (i.e., $\abs{s-t}= l$) share 
the same weight $G_{ts} = w_l$. See Figure~\ref{fig:time-dep0} for an example with 
$\cL=\cbr{1,4}$. Given $\cL$ and $\bw$, the corresponding graph regularizer 
becomes 
\begin{equation}
  \GR{X} = \frac{1}{2} \sum_{l\in\cL} \sum_{t:t>l} w_l \rbr{\bx_t - \bx_{t-l}}^2 + \frac{\eta}{2} \sum_{t} \norm{\bx_t}^2.
  \label{eq:temporal-graph-reg}
\end{equation}

This direct use of graph-based approach for temporal 
dependency, while intuitive, has two issues: a) there might be negatively 
correlated dependencies between two time points; however, the standard 
graph regularization requires nonnegative weights. b) unlike
many applications where such regularizers are used, the 
explicit temporal dependency structure is usually not available and has to be inferred. As a result, 
Most existing temporal MF approaches consider only very simple temporal 
dependencies such with a small size of $\cL$ (e.g., $\cL=\cbr{1}$) and/or a uniform 
weight (e.g., $w_l = 1,\ \forall l\in\cL$). For example, a simple chain graph 
is considered to design the smooth regularizer in TCF~\cite{LX10a}. This leads 
to poor forecasting abilities of existing MF methods for large-scale time 
series applications.

\subsection{Challenges to Learn Temporal Dependencies}
\label{sec:challenge-dep-learning}
One could try to learn the weights $w_{l}$ automatically, by 
using the same regularizer as in \eqref{eq:temporal-graph-reg} but with the 
weights unknown. This would lead to the following optimization problem: 
%
\begin{align}
  \min_{F,X,\bw\ge\b0}\ \sum_{(i,t)\in\Omega} \rbr{Y_{it}- \bff_i^\top \bx_t}^2 + 
  \lambda_f \Rf{F} + \frac{\lambda_x}{2} \sum_{l\in \cL}\sum_{t: t-l > 0} w_l  
  \rbr{\bx_t - \bx_{t-l}}^2 + \frac{\lambda_x\eta}{2} \sum_{t} \norm{\bx_t}^2,
  \label{eq:auto-learning-1}
\end{align}
where $\b0$ is the zero vector, and $\bw\ge\b0$ is the constraint imposed by 
graph regularization.

It is not hard to see that the above optimization yields the trivial all-zero 
solution for $\bw^*$, meaning the objective function is minimized when no 
temporal dependencies exist! 
To avoid the all zero solution, one might want to impose a simplex constraint 
on $\bw$ (i.e., $\sum_{l\in\cL} w_l = 1$). Again, it is not hard to see that 
this will result in $\bw^*$ being a 1-sparse vector, with the non zero 
component $w_{l^*}$ being 1, where 
\[
  l^* = \arg\min_{l\in\cL}\sum_{t:t>l} \| \bx_t - \bx_{t-l} \|^2.
\]
Thus, looking to learn the weights automatically by simply plugging in the 
regularizer in the MF formulation is not a viable solution. 
%

\section{Temporal Regularized Matrix Factorization}
\label{sec:trmf}
In order to resolve the limitations mentioned in Sections~\ref{sec:mf-relates} and \ref{sec:challenge-dep-learning},  
we propose the {\bf T}emporal {\bf R}egularized {\bf M}atrix {\bf F}actorization (TRMF) framework, 
which is a novel approach to incorporate temporal dependencies into matrix 
factorization models. 

We propose to use well-known time series models (e.g. AR) to describe temporal dependencies in $X$ explicitly.  Unlike 
the aforementioned graph-based approaches, there are many well-studied models 
which are specifically designed for data with temporal dependency~\cite{GP09a}. 
Let $M_\Theta$ denote the time-series model for $\cbr{\bx_t}$ as introduced in 
\eqref{eq:dlm-state}. To  incorporate the temporal dependency in TRMF, we 
propose to design a new regularizer $\T{X}$ which   
encourages the structure induced by $M_\Theta$. 

Taking a standard approach to model time series, we set $\T{X}$ 
be the negative log likelihood of observing a particular realization of the 
$\cbr{\bx_t}$ for a given model $M_\Theta$:

\begin{equation}
  \T{X} = - \log \Prob{\bx_1,\ldots,\bx_T\mid \Theta}.
  \label{eq:temporal-reg}
\end{equation}
When $\Theta$ is given, we can use $\Rx{X} = \T{X}$ in the MF 
formulation~\eqref{eq:mf} to encourage $\cbr{\bx_t}$ to follow the temporal 
dependency induced by $M_\Theta$.  When the $\Theta$ is unknown, we can 
treat $\Theta$ as another set of variables and include another regularizer 
$\Rtheta{\Theta}$ into \eqref{eq:mf} as follows.  
\begin{align}
  \min_{F,X,\Theta}\quad \sum_{(i,t) \in \Omega} \rbr{Y_{it} - \fb_i^\top 
  \xb_t}^2 + \lambda_f \Rf{F}+ {\lambda_x \T{X}+\lambda_\theta 
\Rtheta{\Theta}},
  \label{eq:trmf-0}
\end{align}
which be solved by an alternating minimization procedure over $F$, $X$, and $\Theta$.  

\paragraph{Data-driven Temporal Dependency Learning in TRMF:}
Recall that in Section~\ref{sec:challenge-dep-learning}, we showed that 
directly using graph based regularizers to incorporate temporal dependencies 
leads to trivial solutions for the weights. We now show that the TRMF 
framework circumvents this issue.  
%
When $F$ and $X$ are fixed, \eqref{eq:trmf-0} is reduced to the following 
problem: 
\begin{equation}
  \min_{\Theta}\quad \lambda_x \T{X} + \lambda_{\theta} \Rtheta{\Theta},
  \label{eq:map}
\end{equation}
which is a maximum-a-posterior (MAP) estimation problem (in the Bayesian 
sense, assuming priors on the data) to estimate the best $\Theta$ for a given  
$\cbr{\bx_t}$ under the $M_\Theta$ model. There are well-developed algorithms to solve the MAP   
problem~\eqref{eq:map} and obtain non-trivial dependency parameter $\Theta$. 
Thus, unlike most existing temporal matrix factorization approaches where the 
strength of dependencies is fixed, $\Theta$ in TRMF  can be learned 
automatically from data.  

\subsection{Time-Series Analysis with TRMF}
\label{sec:trmf-app}
We can see that TRMF~\eqref{eq:trmf-0} lends itself seamlessly to handle a 
variety of commonly encountered tasks in analyzing data with temporal dependency:
\begin{itemize}
  \item {\bf Time-series Forecasting:} As TRMF comes with a full specification of  
$M_\Theta$ for latent embeddings $\cbr{\bx_t: 1,\ldots,T}$, we can use 
$M_\Theta$ to predict future latent embeddings $\cbr{\bx_t: t > T}$ and have 
the ability to obtain non-trivial forecasting results for $\by_t = F \bx_t$ 
for $t>T$.  
  \item {\bf Missing-value Imputation:} In some time-series applications,  
    certain entries in $Y$ might be unobserved, for example, due to faulty sensors in 
    electricity usage monitoring or occlusions in the case of motion recognition 
    in video. We can use $\bff_i^\top \bx_t$ to impute these missing entries, much 
    like standard matrix completion. This task is useful in many recommender 
    systems~\cite{LX10a} and sensor networks~\cite{YZ09a}. 
\item {\bf Time-series classification/clustering:} The obtained $\bff_i$ can be used 
  as the latent embedding for the $i$-th time series of $Y$. These latent 
  features can be used to perform classification/clustering of the time 
  series.  Note that this can be done even when there are missing entries in the observed  
  data, as missing entries are not a bottleneck for learning $F$.
\end{itemize}

\subsection{Extensions to Incorporate Extra Information}
\label{sec:trmf-ext}
Like matrix factorization approaches, TRMF~\eqref{eq:trmf-0} 
can be extended to incorporate additional information. Below we only briefly 
describe three approaches, and more details on these extensions can be found in Appendix~\ref{sec:ext-details}.

\begin{itemize}
  \item {\bf Known features for time series:} In many applications, one is 
    given additional features along with the observed time series.  
    Specifically, given a set of feature vectors $\cbr{\ba_i\in \RR^{d}}$ for 
    each row of $Y$, we can look to solve  
    \begin{align}
    \notag
      \min_{F,X,\Theta}\quad & \sum_{(i,t) \in \Omega}
      \rbr{Y_{it} - \ba_i^\top F \xb_t}^2 + \lambda_f \Rf{F} \\
      &+ \lambda_x \T{X}+\lambda_\theta 
      \Rtheta{\Theta}. 
    \end{align}

    That is, the observation $Y_{it}$ is posited to be a bilinear function of the 
    feature vector $\ba_i$ and the latent vector $\bx_t$.  Such an inductive 
    framework has two advantages: we can generalize TRMF to a new time series 
    without any observations up to time $T$ (i.e., a new row $i'$ of $Y$ without 
    any observations). As long as the  feature vector $\ba_{i'}$ is available, the 
    model learned by TRMF can be used to estimate 
    $Y_{i' t} = \ba_{i'}^\top F \xb_t,\ \forall t$. Furthermore, prediction can be 
    significantly sped up when $d \ll n$, since the dimension of $F$ is reduced 
    from $n\times k$ to $d\times k$.  Such methods for standard multi-label 
    learning and matrix completion have been previously considered in 
    \cite{PJ13a,MX13a,HFY14b}.  

  \item { \bf Graph information among time series:} Often, separate 
    features for the time series are not known, but other relational 
    information is available. When a graph that encodes pairwise interactions 
    among multiple time series is available, one can incorporate this graph in our 
    framework using the graph regularization approach~\eqref{eq:graph-reg}. 
    Such cases are common in inventory and sales tracking, 
    where sales of one item is related to sales of other items. Given a graph 
    $G^f$ describing the relationship among multiple time series, we can 
    formulate a graph regularized problem: 
    \begin{align}
    \notag
      \min_{F,X,\Theta}\quad & \sum_{(i,t) \in \Omega} \rbr{Y_{it} - \bff_i^\top  
      \bx_t}^2 + \lambda_f \GR[f]{F} \\ 
      &+ \lambda_x \T{X}+\lambda_\theta \Rtheta{\Theta},
    \end{align}
    where $\GR[f]{F}$ is the graph regularizer defined in~\eqref{eq:graph-reg} 
    capturing pairwise interactions between time series.  Graph regularized matrix 
    completion methods have been previously considered in \cite{TZ12a,NR15a}.

  \item { \bf Temporal-regularized tensor factorization:}  Naturally, 
    TRMF can be easily extended to analyze temporal collaborative filtering 
    applications~\cite{LX10a,JS12a}, where the targeted data is a tensor with certain modes evolving over time. For example, consider $\cY\in\RR^{m\times n\times T}$ be a  
    3-way tensor with $Y_{ijt}$ encoding the rating of the $i$-th user for the 
    $j$-th item at time point $t$. We can consider the following temporal 
    regularization tensor factorization (TRTF) with $\T{X}$ as follows: 
    \begin{align}
    \notag
      \min_{P,Q,X,\Theta}\quad & \sum_{(i,j,t)\in\Omega} \rbr{Y_{ijt} - 
      \abr{\bp_i,\bq_j,\bx_t}}^2 + \lambda_p \cR_p(P) \\
      & + \cR_q(Q) + \T{X} + \Rtheta{\Theta},
    \end{align}
    where $P=[\bp_1,\cdots,\bp_m]^\top \in \RR^{m\times k}$ and 
    $Q=[\bq_1,\cdots,\bq_n]^\top \in \RR^{n\times k}$ are the 
    latent embeddings for the $m$ users and $n$ items, respectively, and with some abuse of notation, we define
    $
      \abr{\bp_i,\bq_j,\bx_t} = \sum_{r=1}^k p_{ir} q_{jr} x_{tr}.
    $
     
\end{itemize}

\section{A Novel Autoregressive Temporal Regularizer}
\label{sec:armf}

Up until now, we described the TRMF framework in a very general sense, with 
the regularizer $\T{X}$ incorporating  dependencies specified by the time 
series model $M_{\Theta}$. In this section, we specialize this to the case of 
AR models.

An AR model, parameterized by a lag set $\cL$ and weights
$\cW=\cbr{W^{(l)} \in \RR^{k\times k}:l\in\cL}$, assumes that $\bx_t$ is a 
noisy linear combination of some previous points; that is, 
\begin{equation}
  \bx_t = \sum_{l\in\cL} W^{(l)} \bx_{t-l} + \bepsilon_t,
  \label{eq:ar}
\end{equation}
where $\bepsilon_t$ is Gaussian noise vector. For simplicity, we assume that the 
$\bepsilon_t\sim \cN(0, \sigma^2 I_k)$.\footnote{If the (known) covariance 
matrix is not identity, we can suitably modify the regularizer.} As a result, 
the  
temporal regularizer $\T{X}$ corresponding to this AR model can be written as 
follows: 
\begin{align}
  \AR{X} := \sum_{t=m}^T \frac{1}{2}\norm{ \bx_t - \sum_{l \in \cL} W^{\rbr{l}} 
  \bx_{t-l}}^2 + \frac{\eta}{2}\sum_{t}\norm{\bx_t}^2,
  \label{eq:tar}
\end{align}
where $m := 1+L$, $L := \max\rbr{\cL}$, and the last term with a positive 
$\eta$ is used to guarantee the strong convexity of \eqref{eq:tar}. We denote 
the term in \eqref{eq:tar} by $\AR{X}$. 

TRMF allows us to learn the weights $\cbr{W^{(l)}}$ when they are unknown. Since  
each $W^{(l)} \in \RR^{k \times k}$, there will be $\abs{\cL} k^2$ 
variables to learn, which may lead to overfitting. To prevent this and to 
yield more interpretable results, we consider  
diagonal $W^{(l)}$, which reduces 
the number of parameters to $\abs{\cL} k$. To simplify notation, we 
use $\cW$ to denote the $k\times L$ matrix with the $l$-th 
column constituting the diagonal elements of $W^{(l)}$. Note 
that for $l\notin\cL$, the $l$-th column of $\cW$ is a zero vector. 
Let $\bxbar_r^\top = [\cdots,X_{rt},\cdots]$ be the $r$-th row of $X$ and 
$\bwbar_r^\top = [\cdots,\cW_{rl},\cdots]$ be the $r$-th 
row of $\cW$. Then \eqref{eq:tar} can be written as follows:
\begin{align}
  \AR{X} &= \sum_{r=1}^k \ARr[r]{\bxbar_r}, \\
  \ARr{\bxbar} &= \frac{1}{2}{\sum_{t=m}^T \rbr{x_{t} - \sum_{l\in\cL} w_{l} 
  x_{t-l}}^2}+ \frac{\eta}{2}\norm{\bxbar}^2,
  \label{eq:tar-r}
\end{align}
where $x_t$ is the $t$-th element of $\bxbar$, and $w_l$ is the $l$-th 
element of $\bwbar$. 

{\bf Correlations among Multiple Time Series.} 
TRMF retains the power to capture the correlations among time series 
via the factors $\cbr{\bff_i}$, even after simplifying $\cbr{W^l}$ to be diagonal,
since it
 has effects only on the structure of latent embeddings 
$\cbr{\bx_t}$. 
Indeed, as the $i$-th dimension of $\cbr{\by_t}$ is 
modeled by $\bff_i^\top X$ in~\eqref{eq:trmf-0}, one can see the low rank $F$ 
as a $k$ dimensional latent embedding of multiple time series. This low rank 
embedding captures correlations among multiple time series. Furthermore, 
$\cbr{\bff_i}$ acts as time series features, 
which can be used to perform classification/clustering even in the 
presence of missing values.

{\bf Choice of Lag Index Set $\cL$.}  Unlike most MF approaches mentioned 
in Section~\ref{sec:mf-relates}, the choice of $\cL$ in TRMF is more flexible. 
Thus, TRMF can provide the following two important advantages for 
practitioners: First, because there is no need to specify the weight parameters 
$\cW$, $\cL$ can be chosen to be larger to account for long range 
dependencies, which also yields more accurate and robust forecasts. Second, 
the indices in $\cL$ can be discontinuous so that one can easily 
embed {\em domain knowledge} about periodicity or seasonality.
For example, one might consider 
$\cL = \cbr{1,2,3,51,52,53}$ for weekly data with an one-year seasonality.

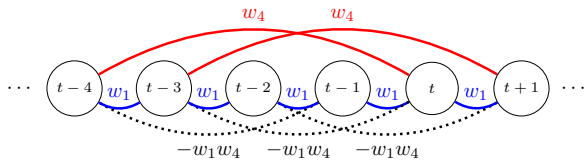
\begin{figure}
  \centering
  \begin{resize}{0.48\linewidth}
    \begin{tikzpicture}
      \node [draw,minimum width=2.5em,minimum height=2.5em, circle] at (0,0) (x0) {\scriptsize{$t$}};
      \foreach \t in {-1,-2,-3,-4,+1} {
        \node [draw,minimum width=2.5em,minimum height=2.5em, circle] at (1.5*\t, 0) (x\t)
        {\scriptsize{$t\t$}};
      }
      \node at (1.5*-4.6,0) {$\cdots$};
      \node at (1.5*1.6,0) {$\cdots$};
      \path (x0) edge [draw=blue,very thick,bend left=30] node [above] {\color{blue}$w_1$} (x-1);
      \path (x0) edge [draw=red,very thick,bend right=30] node [above] {\color{red}$w_4$} (x-4);
      \path (x0) edge [draw=black,dotted, very thick,bend left=30] node [below] {\color{black}$-w_1w_4$} (x-3);
      \path (x-1) edge [draw=black,dotted, very thick,bend left=30] node [below] {\color{black}$-w_1w_4$} (x-4);
      \path (x+1) edge [draw=black,dotted, very thick,bend left=30] node [below] {\color{black}$-w_1w_4$} (x-2);
      \path (x+1) edge [draw=blue,very thick,bend left=30] node [above] {\color{blue}$w_1$} (x0);
      \path (x+1) edge [draw=red,very thick,bend right=30] node [above] {\color{red}$w_4$} (x-3);
      \path (x-1) edge [draw=blue,very thick,bend left=30] node [above] {\color{blue}$w_1$} (x-2);
      \path (x-2) edge [draw=blue,very thick,bend left=30] node [above] {\color{blue}$w_1$} (x-3);
      \path (x-3) edge [draw=blue,very thick,bend left=30] node [above] {\color{blue}$w_1$} (x-4);
    \end{tikzpicture}
  \end{resize}
  \caption{Graph structure induced by the AR temporal 
  regularizer~\eqref{eq:tar-r} with $\cL = \cbr{1,4}$. }
  \label{fig:ar-graph}
\end{figure}

{\bf Connections to Graph Regularization.} 
We now establish connections between $\ARr{\bxbar}$ and graph regularization~\eqref{eq:graph-reg} for matrix factorization. 
%
%
 Let 
$\cLbar:= \cL \cup \{0\}$,  
$w_0 = -1$ so that \eqref{eq:tar-r} can be written as 
\[
  \ARr{\bxbar} = \frac{1}{2}\sum_{t=m}^T \rbr{\sum_{l\in \cLbar} w_{l} 
  x_{t-l}}^2 + \frac{\eta}{2}\norm{\bxbar}^2,
\]
and  
let $\delta(d):= \cbr{l\in\cLbar: l-d\in\cLbar}$. We then have 
the following result: 
\begin{theorem}
\label{thm:ar-graph}
Given a lag index set $\cL$, weight vector $\bwbar\in \RR^{L}$, and 
$\bxbar \in \RR^T$, there is a weighted signed graph $\GAR$ with $T$ nodes and a 
diagonal  matrix $D\in \RR^{T\times T}$ such that  
  \begin{align}
    \ARr{\bxbar} = \GR[\text{AR}]{\bxbar}+\frac{1}{2}\bxbar^\top D \bxbar,
    \label{eq:ar-graph}
  \end{align}
  where $\GR[\text{AR}]{\bxbar}$ is the graph 
  regularization~\eqref{eq:graph-reg} for $\GAR$.  
  The edge weight for $\GAR$ and the diagonal element of $D$ are described as follows. 
  \begin{align*}
    \GAR_{t,t+d} &= \begin{cases}
      \displaystyle\sum_{\substack{l\in\delta(d)}}\sum_{m\le t+l\le T} 
      -w_lw_{l-d} & \text{ if }\ \delta(d) \neq \phi, \\
      0  & \text{ otherwise,}\\
    \end{cases}\quad \forall t, d, \\
    D_{tt} &= \rbr{\sum_{l\in\cLbar} w_l}\rbr{\sum_{\substack{l\in \cLbar}} 
    w_l[m\le t+l \le T]}\quad \forall t.
  \end{align*}
\end{theorem}
See Appendix \ref{sec:proof-ar-graph} for the detailed proof of Theorem \ref{thm:ar-graph}.
From Theorem~\ref{thm:ar-graph}, we know that $\delta(d)$ is not empty if and 
only if there are edges of distance $d$ in $\GAR$. Thus, we can construct the 
dependency graph for a $\ARr{\bxbar}$ by checking whether $\delta(d)$ is 
empty. Figure~\ref{fig:ar-graph} demonstrates an example with 
$\cL = \cbr{1,4}$. We can see that in addition to the edges of distance $d=1$ 
and $d=4$, there are also edges of distance $d=3$ (dotted edges in 
Figure~\ref{fig:ar-graph}) because $4-3\in \cLbar$ and  
$\delta(3) = \cbr{4}$. 

Although Theorem~\ref{thm:ar-graph} shows that AR-based regularizers are
similar to the graph-based regularization framework, we note the following key 
differences:  
\begin{itemize}
 \item The graph $\GAR$ defined in Theorem~\ref{thm:ar-graph} is a signed 
    weighted graph, which contains both positive and negative edges. This 
    implies that the AR temporal regularizer is able to support negative 
    correlations, which the standard graph-based regularizer does not. This 
    can make $\GR[\text{AR}]{\bxbar}$ non-convex.  
  \item While $\GR[\text{AR}]{\bxbar}$ might be non-convex, the addition of 
    the second term in \eqref{eq:ar-graph} still leads to a convex regularizer 
    $\ARr{\bxbar}$.
  \item Unlike the approach~\eqref{eq:temporal-graph-reg} where there is 
    freedom to specify a weight for each distance, in the graph defined in 
    Theorem~\ref{thm:ar-graph}, the weight values for the edges are more 
    structured (e.g., the weight for $d=3$ in Figure~\ref{fig:ar-graph} is 
    $-w_1w_4$).  Hence, minimization w.r.t. $w's$ is not trivial, and neither 
    are the obtained solutions.  
\end{itemize}

\subsection{Optimization for TRMF with AR Temporal Regularization} 
Plugging $\T{X} = \AR{X}$ into \eqref{eq:trmf-0}, we obtain the following  
optimization problem: 
\begin{align}
  \min_{F,X,\cW}\quad \sum_{(i,t) \in \Omega} \rbr{Y_{it} - \bff_i^\top 
  \bx_t}^2 + \lambda_f \Rf{F} + \sum_{r=1}^k \lambda_x \ARr[r]{\bxbar_r} + 
  \lambda_w \Rw{\cW},
  \label{eq:trmf-ar}
\end{align}
where $\Rw{\cW}$ is a regularizer for $\cW$. We will refer to ~\eqref{eq:trmf-ar} as TRMF-AR.
We can apply alternating minimization to solve 
\eqref{eq:trmf-ar}, and  in fact, solving for each variable reduces to well known methods, for which highly efficient algorithms exist:

%

{\bf Updates for $F$.} When $X$ and $\cW$ are fixed, the subproblem of 
updating $F$ is the same as updating $F$ while $X$ fixed in 
\eqref{eq:mf}.  Thus, fast algorithms such as alternating least 
squares~\cite{YK09a} or coordinate descent~\cite{HFY14a} can be applied 
directly to find $F$.  

{\bf Updates for $X$.} To update $X$ when $F$ and $\cW$ fixed, we solve:
\begin{align}
  \arg\min_{X}\quad \sum_{(i,t)\in\Omega} \rbr{Y_{it}-\bff_i^\top\bx_t}^2 + 
  \lambda_x \sum_{r=1}^k \ARr[r]{\bxbar_r}. 
  \label{eq:updatex}
\end{align}
From Theorem~\ref{thm:ar-graph}, we see that $\ARr{\bxbar}$ shares the same 
form as the graph regularizer, and we can apply GRALS~\cite{NR15a} to solve \eqref{eq:updatex}.
%
%

{\bf Updates for $\cW$.}
How to update $\cW$ while $F$ and $X$ fixed depends on the choice of $\Rw{\cW}$.
There are many parameter estimation techniques developed for AR with various 
regularizers~\cite{HW07a,WN14a}. For simplicity, we 
consider the squared Frobenius norm: $\Rw{\cW} = \norm{\cW}_{F}^2$.  As a 
result, each row of $\bwbar_r$ of $\cW$ can be updated independently by  
solving the following one-dimensional autoregressive problem. 
\begin{align}
  &\arg\min_{\bwbar}\quad \lambda_x \ARr{\bxbar_r} + \lambda_w \norm{\bwbar}^2 
  \label{eq:updatew} \\
  = &\arg\min_{\bwbar}\quad \sum_{t=m}^T \rbr{x_t - \sum_{l\in\cL} w_l 
  x_{t-l}}^2 + \frac{\lambda_w}{\lambda_x} \norm{\bwbar}^2. \label{eq:updatew-2}
\end{align}
\eqref{eq:updatew-2} is  a simple $\abs{\cL}$ dimensional 
ridge regression problem with $T-m+1$ instances, which can be solved 
efficiently by Cholesky factorization such as the backslash operator in 
MATLAB. 

Note that since out method is highly modular, one can resort to {\em any} 
method to solve the optimization subproblems that arise for each module. 
Moreover, as seen from Section \ref{sec:trmf-ext}, TRMF can also be used with 
different regularization structures, making it highly adaptable.

\subsection{Connections to Existing MF Approaches}
\label{sec:armf-existing}
TRMF-AR, while being a special family of TRMF, still can be seen as a 
generalization for many existing MF approaches to handle data with temporal 
dependencies. We demonstrate several cases  which arise from a 
specific choice of  $\cL$ and $\cbr{W^{(l)}}$:
\begin{itemize}
 \item Temporal Collaborative Filtering~\cite{LX10a}: AR(1) with $W^{\rbr{1}} = I_k$ 
    on $\cbr{\xb_t}$.  
  \item NMF with Temporal Smoothness~\cite{ZC05a}:  AR($L$) with 
    $W^{\rbr{l}} = \alpha^{l-1}(1-\alpha)I_k,\ \forall l=1,\ldots,L$, where $\alpha$ is a 
    pre-defined parameter.   
  \item Sparsity Regularized Matrix Factorization~\cite{YZ09a,MR12a}: AR(1) 
    with $W^{\rbr{1}} = I_n$ on $\cbr{F\bx_t}$.  
  \item Temporal stability in low-rank structure~\cite{SR10a}: AR(2) with 
    $W^{\rbr{1}} = 2I_n$ and $W^{\rbr{2}} = -I_n$ on $\cbr{F\bx_t}$. This is 
    also corresponds to the Hodrick-Prescott filter~\cite{RH97a,SJK09a}. 
  \item Dynamic Linear Model~\cite{RK60a}: DLM is a time 
    series model which can be directly applied to model $Y$. It is essentially  
    a latent AR(1) model with a general $W^{\rbr{1}}$, which can be estimated 
    by expectation-maximization algorithms~\cite{RS82a,ZG96a,LL09a}.    
\end{itemize}

\subsection{Connections to Gaussian Markov Random Field} 
\label{sec:armf-invcov}
The Gaussian Markov Random Field (GMRF) is 
a general way to model multivariate data with dependencies. GMRF assumes that 
data are generated from a multivariate  
Gaussian distribution with a covariance matrix $\Sigma$ which describes the 
dependencies among $T$ dimensional variables i.e., $\bxbar \sim \cN(0, \Sigma)$.\footnote{a 
non zero mean can be similarly handled.}   
If the unknown $\bxbar$ is assumed to 
be generated from this model,  The negative log likelihood of the data can be written as
$$  \bxbar^\top \Sigma^{-1} \bxbar, $$
ignoring the constants and where $\Sigma^{-1}$ is the inverse covariance for the Gaussian distribution. This prior can be incorporated into an empirical 
risk minimization framework as a regularizer. Furthermore, it is known that
 if 
$\rbr{\Sigma^{-1}}_{st} = 0$, then $x_t$ and $x_s$ are conditionally 
independent, given the other variables. In 
Theorem~\ref{thm:ar-graph} we established connections to graph based 
regularizers, and that such methods can be seen as regularizing with the 
inverse covariance matrix for Gaussians~\cite{TZ12a}. We thus have the following 
result: 
 
\begin{corollary}
  For any lag set $\cL$, $\bwbar$, and $\eta>0$, the inverse covariance 
  matrix $\Sigma^{-1}$ of the GMRF model corresponding to the quadratic regularizer 
    $\Rx{\bxbar} := \ARr{\bxbar}$
  shares the same off-diagonal non-zero pattern as $\GAR$ defined in 
  Theorem~\ref{thm:ar-graph} for $\ARr{\bxbar}$.
  \label{thm:ar-invcov}
\end{corollary}
A detailed proof is in Appendix~\ref{sec:proof-ar-invcov}. As a result, our proposed AR-based regularizer is 
equivalent to imposing a Gaussian prior on $\bxbar$ with a structured inverse 
covariance described by the matrix $\GAR$ defined in Theorem~\ref{thm:ar-graph}. 
Moreover, the step to learn $\cW$ has a natural interpretation: the lag set 
$\cL$ imposes the non-zero pattern of the graphical model on the data, and then we solve a 
simple least squares problem to learn the weights corresponding to the edges.

\section{Experimental Results}
\label{sec:exp}

In this section, we perform extensive experiments on time series forecasting 
and missing value imputations on both synthetic and real-world datasets. 

{\bf Datasets Used:}
\begin{itemize}
  \item \trmfsyn: a small synthetic dataset with $n=16, T=128$. We generate $\cbr{\bx_t\in\RR^{4}: t=1,\ldots,128}$ from the 
    autoregressive process \eqref{eq:ar} with a lag index set 
    $\Lcal = \cbr{1,8}$, randomly generated $\cbr{W^{(l)}}$, and an additive 
    white Gaussian noise of $\sigma = 0.1$. We then randomly generate a matrix 
    $F\in\RR^{16\times 4}$ and obtain $\by_t = F\bx_t + \epsilon$, where 
    $\epsilon \sim \mathcal{N}(0,0.1)$. 
    \item \elec\footnote{\url{https://archive.ics.uci.edu/ml/datasets/ElectricityLoadDiagrams20112014}.}: 
      the electricity usage in kW recorded every 15 minutes, for $n = 370$ clients. We convert the data to reflect hourly consumption, by aggregating blocks of 4 columns, to obtain $T = 26,304$.
  \item \traffic\footnote{\url{https://archive.ics.uci.edu/ml/datasets/PEMS-SF}.}:
    A collection of 15 months of daily data from the California Department of Transportation. The data describes the occupancy
rate, between 0 and 1, of different car lanes of San Francisco bay area freeways. The data was sampled every 10 minutes, and we again aggregate the columns to obtain hourly traffic data to finally get $n=963$, 
    $T=10,560$. 
  \item \wmone \& \wmtwo: two propriety datasets from Walmart E-commerce contain weekly sale information
  of 1,350 and 1,582 items for 187 weeks, respectively. The time-series of sales 
  for each item start and end at different time points; for modeling purposes we 
  assume one start and end timestamp by padding each series with missing values. 
  This along with some other missing values due to out-of-stock reasons lead to 
  55.3\% and 49.3\% of entries being missing.  
\end{itemize}

{\bf Compared Methods/Implementations:}
\begin{itemize}
  \item {TRMF-AR:} The proposed formulation~\eqref{eq:trmf-ar} with 
      $\Rw{\cW}=\norm{\cW}_F^2$. For the lag set $\cL$, we use 
      $\cbr{1,2,\ldots,8}$ for \trmfsyn, 
      $\cbr{1,\ldots,24} \cup \cbr{7\times 24,\ldots,8\times 24-1}$ for \elec 
      and \traffic, and $\cbr{1,\ldots,10}\cup\cbr{50,\ldots,56}$ for \wmone 
      and \wmtwo. 
  \item {SVD-AR(1):} The best rank-$k$ approximation of $Y = U S V^\top$ is first 
    obtained by singular value decomposition. After setting $F=US$ and $X=V^\top$, a $k$-dimensional AR(1) is 
    learned on $X$ for forecasting.   
  \item {TCF:} Matrix factorization with the simple temporal regularizer 
    proposed in \cite{LX10a}.   
  \item {AR(1):} $n$-dimensional AR(1) model. 
  \item {DLM:} We tried two DLM implementations: the R-DLM 
    package~\cite{GP10a} which is widely used in  
    the time-series community and the DLM, the code provided in 
    \cite{LL11a}.
  \item {Mean:} This is the baseline approach, which predicts everything to be 
    the mean of the observed portion of $Y$. 
\end{itemize}

{\bf Evaluation Criteria:} As the range of values varies in different time 
series data sets, we compute two normalized criteria: normalized 
deviation (ND) and normalized RMSE (NRMSE) as follows. 
\begin{align*}
  \text{Normalized deviation (ND):} &\quad
  {\rbr{\frac{1}{\abs{\Omega_{test}}}\displaystyle\sum_{(i,t)\in\Omega_{test}}\abs{\Yhat_{it} - Y_{it}}}}
  \Bigg/\rbr{\frac{1}{\abs{\Omega_{test}}}\sum_{(i,t)\in\Omega_{test}} \abs{Y_{ij}}} \\
  \text{Normalized RMSE (NRMSE):}   &\quad
  {\sqrt{\frac{1}{\abs{\Omega_{test}}}\displaystyle\sum_{(i,t)\in\Omega_{test}}\rbr{\Yhat_{it} - Y_{it}}^2}}
  \Bigg/\rbr{\frac{1}{\abs{\Omega_{test}}}\sum_{(i,t)\in\Omega_{test}} \abs{Y_{ij}}}
\end{align*}
For each method and data set, we perform the grid search over various parameters (such as 
$k$, $\lambda$ values) and report the best ND and NRMSE.  We search 
$k \in \{2,4,8\}$ for synthetic and $\in \{20,40\}$ for other datasets. For  
TRMF-AR, SVD-AR(1), TCF, and AR(1), we search $\lambda \in \{50,5,0.5,0.05\}$

\begin{table*}[t]
  \centering
  \caption{Forecasting results: ND/ NRMSE for 
    each approach.
    Lower values are better. 
    ``-'' indicates an unavailability due to scalability or an inability to 
    handle missing values. 
    DLM denotes the implementation provided in ~\cite{LL11a}, 
    while R-DLM denotes the popular DLM implementation in R~\cite{GP10a}. Lower values are better. 
    ``-'' denotes that the result is not available because the implementation 
    cannot scale up to size the data or deal with the scenario of missing 
    values.  TRMF-AR, which is the proposed approach~\eqref{eq:trmf-ar}, 
  outperforms all other considered approaches in most situations.} 
  \label{tab:forecast}
    \begin{tabular}{@{}lccc@{ }||cccc@{}}
      \multicolumn{8}{c}{Forecasting with Full Observation} \\
      & \multicolumn{3}{c||}{Matrix Factorization Models} & \multicolumn{4}{c}{Time Series 
      Models}\\
      & TRMF-AR& SVD-AR(1)& TCF& AR(1)& DLM&  R-DLM & Mean \\
      \hline
      \hline
      \trmfsyn & \best{0.373}/ \best{0.487} &  {0.444}/ {0.872} &  {1.000}/ {1.424} &      {0.928}/ {1.401} & {0.936}/ {1.391} & {0.996}/ {1.420}  &{1.000}/ {1.424} \\
      \elec    &      {0.255}/ \best{1.397} &  {0.257}/ {1.865} &  {0.349}/ {1.838} & \best{0.219}/ {1.439} & {0.435}/ {2.753} & {-}/ {-}          & {1.410}/ {4.528} \\
      \traffic & \best{0.185}/ \best{0.421} &  {0.555}/ {1.194} &  {0.624}/ {0.931} &      {0.275}/ {0.536} & {0.639}/ {0.951} & {-}/ {-}          & {0.560}/ {0.826}\\
      \hline
      \hline
      \multicolumn{8}{c}{Forecasting with Missing Values} \\
      \wmone   & \best{0.533}/ \best{1.958} &  {-}/ {-}         & {0.540}/{2.231}   & {-}/ {-}              & {0.602}/ {2.293} & {-}/ {-}          & {1.239}/{3.103} \\
      \wmtwo   & \best{0.432}/ \best{1.065} &  {-}/ {-}         & {0.446}/{1.124}   & {-}/ {-}              & {0.453}/ {1.110} & {-}/ {-}          & {1.097}/{2.088}
    \end{tabular}
\end{table*}
\subsection{Forecasting} 
We compare the forecasting 
performance of various approaches. 
The detailed results are shown in Table~\ref{tab:forecast}. 

{\bf Remark on the implementations of DLM.} 
We first note that the numbers for R-DLM on \elec and \traffic are not available because the R 
package crashes when the dimension of the time series is large (See 
Appendix~\ref{sec:rdlm} for the source code to demonstrate that R-DLM fails when $n=32$). 
Thus, the only numbers we obtained for R-DLM are on \trmfsyn. Furthermore, the 
\verb!dlmMLE! routine in R-DLM uses a general optimization solver,\footnote{It 
uses LBFGS-B with a finite-difference approximation to obtain gradients.} 
which is orders of magnitude slower than the DLM implementation provided in \cite{LL11a}. 

{\bf Forecasting with Full Observations.} We first compare various methods on 
the task of forecasting values in the test set, given fully observed training 
data.  Three data sets are considered: \trmfsyn, \elec, and \traffic.
For \trmfsyn, we consider one-point ahead forecasting task and use the 
last ten time points as the test periods. For \elec and \traffic, 
we consider the 24-hour ahead forecasting task and use last seven days 
as the test periods.  The results are shown in the first part of 
Table~\ref{tab:forecast}. We can clearly observe  
the superiority of TRMF-AR. Other than the ND for \elec, TRMF-AR outperforms 
other existing matrix factorization approaches and 
classic time-series approaches.

{\bf Forecasting with Missing Values.}  We next compare the methods on the 
task of forecasting in the presence of missing values in the training data. We 
use the Walmart datasets here, and consider 6-week ahead forecasting and use  
last 54 weeks as the test periods. Note that SVD-AR(1) and AR(1) cannot handle 
missing values. The second part of Table~\ref{tab:forecast} shows that we 
again outperform other methods.

\begin{table}[t]
  \centering
  \caption{Missing value imputation on time series data sets. Note that TRMF outperforms all competing methods in almost all cases. When sufficiently large amount of data is available in the case of the \elec dataset, DLM is slightly better. } 
  \begin{tabular}{@{}lcccc||cc@{}}
    & \multirow{2}{*}{$\frac{\abs{\Omega}}{n\times T}$} & \multicolumn{3}{c||}{Matrix Factorization Models} & \multicolumn{2}{c}{Time Series 
    Models}\\
    & & TRMF-AR& TCF& MF & DLM& Mean \\
    \hline
    \hline
\multirow{4}{*}{\trmfsyn} 
&20\%&  \best{0.467}/ \best{0.661}&  {0.713}/ {1.030}&  {0.688}/ {1.064}&  {0.933}/ {1.382}&  {1.002}/ {1.474} \\
&30\%&  \best{0.336}/ \best{0.455}&  {0.629}/ {0.961}&  {0.595}/ {0.926}&  {0.913}/ {1.324}&  {1.004}/ {1.445} \\
&40\%&  \best{0.231}/ \best{0.306}&  {0.495}/ {0.771}&  {0.374}/ {0.548}&  {0.834}/ {1.259}&  {1.002}/ {1.479} \\
&50\%&  \best{0.201}/ \best{0.270}&  {0.289}/ {0.464}&  {0.317}/ {0.477}&  {0.772}/ {1.186}&  {1.001}/ {1.498} \\
\hline
\multirow{4}{*}{\elec} 
&20\%&  \best{0.245}/ \best{2.395}&  {0.255}/ {2.427}&  {0.362}/ {2.903}&  {0.462}/ {4.777}&  {1.333}/ {6.031} \\
&30\%&  \best{0.235}/ \best{2.415}&  {0.245}/ {2.436}&  {0.355}/ {2.766}&  {0.410}/ {6.605}&  {1.320}/ {6.050} \\
&40\%&       {0.231}/      {2.429}&  {0.242}/ {2.457}&  {0.348}/ {2.697}&  \best{0.196}/ \best{2.151}&  {1.322}/ {6.030} \\
&50\%&       {0.223}/      {2.434}&  {0.233}/ {2.459}&  {0.319}/ {2.623}&  \best{0.158}/ \best{1.590}&  {1.320}/ {6.109} \\
\hline
\multirow{4}{*}{\traffic} 
&20\%&  \best{0.190}/ \best{0.427}&  {0.208}/ {0.448}&  {0.310}/ {0.604}&  {0.353}/ {0.603}&  {0.578}/ {0.857} \\
&30\%&  \best{0.186}/ \best{0.419}&  {0.199}/ {0.432}&  {0.299}/ {0.581}&  {0.286}/ {0.518}&  {0.578}/ {0.856} \\
&40\%&  \best{0.185}/ \best{0.416}&  {0.198}/ {0.428}&  {0.292}/ {0.568}&  {0.251}/ {0.476}&  {0.578}/ {0.857} \\
&50\%&  \best{0.184}/ \best{0.415}&  {0.193}/ {0.422}&  {0.251}/ {0.510}&  {0.224}/ {0.447}&  {0.578}/ {0.857} \\
  \end{tabular}
\end{table}
\subsection{Missing Value Imputation} 

We next consider the case if imputing missing values in the data.  As 
considered in \cite{LL09a}, we assume that entire blocks of data are missing. 
This corresponds to sensor malfunctions for example, over a length of time.  

To create data with missing entries, we first fixed the percentage of data 
that we were interested in observing, and then uniformly at random occluded 
blocks of a predetermined length ($2$ for synthetic data and $5$ for the real 
datasets). The goal was to predict the occluded values. Table 
\ref{tab:imputation} shows that TRMF outperforms the methods we compared to on 
almost all cases.

\section{Conclusions}
\label{sec:conclusion}
In this paper, we have proposed introduced a novel temporal regularized matrix factorization 
framework (TRMF) for large-scale multiple time series problems with missing values.  
TRMF not only models temporal dependency among the data points, but also 
supports data-driven dependency learning. Our method generalizes several well 
known methods, and also yields superior performance when compared to other 
state-of-the-art methods on real-world datasets.

\section*{Acknowledgments}
This research was supported by NSF grant CCF-1320746 and gifts from Walmart 
Labs and Adobe. We thank Abhay Jha for the help to the experiments on the 
real-world datasets from Walmart E-commerce. 

\small
\bibliographystyle{plain}
\bibliography{trmf-bib}

\appendix

\section{Proofs}
\label{sec:proof}
\subsection{Proof of Theorem~\ref{thm:ar-graph}}
\label{sec:proof-ar-graph}
\begin{proof}
  In this proof, we use the notations and summation manipulation techniques introduced by Knuth~\cite{RG94a}. To prove \eqref{eq:ar-graph}, it suffices to prove that 
  \begin{equation}
    \sum_{m \le t \le T}\rbr{\sum_{l\in\Lcalbar} w_{l} x_{t-l}}^2 = \sum_{1\le 
    t\le T} \sum_{1\le d \le L} \GAR_{t,t+d} \rbr{x_{t}-x_{t+d}}^2 + 
    \bxbar^\top D \bxbar.
    \label{eq:proof-eq}
  \end{equation}
  The LHS of the $\eqref{eq:proof-eq}$ can be expanded and regrouped as follows.
\begin{align*}
  &\sum_{m \le t \le T}\rbr{\sum_{l\in\Lcalbar} w_{l} x_{t-l}}^2\\
  = &\sum_{m\le t \le T} \rbr{{\sum_{l\in\Lcalbar} w_l^2 x_{t-l}^2} + 
  \sum_{1\le d \le L} \sum_{l\in\delta(d)} 2 w_{l}w_{l-d}x_{t-l}x_{t-l+d}} \\
  = &\sum_{m\le t \le T} \rbr{{\sum_{l\in\Lcalbar} w_l^2 x_{t-l}^2}+
  \sum_{1\le d \le L} \sum_{l\in\delta(d)} \rbr{ 
  -w_{l}w_{l-d}\rbr{x_{t-l}-x_{t-l+d}}^2 + w_{l}w_{l-d}\rbr{x_{t-l}^2 + x_{t-l+d}^2}}}\\
  =& \underbrace{\sum_{m\le t \le T}\sum_{1\le d \le L}\sum_{l\in\delta(d)} - 
  w_{l} w_{l-d} \rbr{x_{t-l} - x_{t-l+d}}^2}_{\Gcal(\bxbar)} + 
  \underbrace{\sum_{m\le t \le 
  T}\rbr{\sum_{l\in\Lcalbar} w_l^2 x_{t-l}^2 + \sum_{1\le d\le 
  L}\sum_{l\in\delta(d)} w_{l}w_{l-d}\rbr{x_{t-l}^2 + x_{t-l+d}^2}}}_{\Dcal(\bxbar)}
\end{align*}

Let's look at the first term $\Gcal(\bxbar)$:
\begin{align*}
  \Gcal(\bxbar) &= \sum_{1\le d\le L} \sum_{l\in\delta(d)} \sum_{m\le t \le T} - 
  w_{l} w_{l-d} \rbr{x_{t-l} - x_{t-l+d}}^2 \\
  &= \sum_{1\le d\le L} \sum_{l\in\delta(d)} \sum_{m-l\le t \le T-l} - 
  w_{l} w_{l-d} \rbr{x_{t} - x_{t+d}}^2 \\
  &= \sum_{1\le d\le L} \sum_{l\in\delta(d)} \sum_{1\le t \le T} - 
  w_{l} w_{l-d} \rbr{x_{t} - x_{t+d}}^2  [m-l\le t \le T-l]\\
  &= \sum_{1\le t \le T} \sum_{1\le d\le L} \rbr{\sum_{l\in\delta(d)}  - 
  w_{l} w_{l-d}[m-l\le t \le T-l] }\rbr{x_{t} - x_{t+d}}^2  \\
  &= \sum_{1\le t \le T} \sum_{1\le d\le L} 
  \underbrace{\rbr{\sum_{\substack{l\in\delta(d)\\ m\le t+l\le T}}  - 
  w_{l} w_{l-d}}}_{G_{t,t+d}}\rbr{x_{t} - x_{t+d}}^2,
\end{align*}
where we can see that $\Gcal(\bxbar)$ is equivalent to the first term of RHS 
of \eqref{eq:proof-eq}.

Now, we consider the second term $\Dcal(\bxbar)$:
\begin{align*}
\Dcal(\bxbar) &= \sum_{m\le t \le 
  T}\rbr{\sum_{l\in\Lcalbar} w_l^2 x_{t-l}^2 + \sum_{1\le d\le L}\sum_{l\in\delta(d)} w_{l}w_{l-d}\rbr{x_{t-l}^2 + x_{t-l+d}^2}}\\
  &=\underbrace{\sum_{m\le t \le T}\sum_{l\in\Lcalbar} w_l^2x_{t-l}^2}_{\Dcal_1(\bxbar)}
  +\underbrace{\sum_{m\le t \le T}\sum_{1\le d\le L}\sum_{l\in\delta(d)} w_{l}w_{l-d} x_{t-l}^2}_{\Dcal_2(\bxbar)}
  +\underbrace{\sum_{m\le t \le T}\sum_{1\le d\le L}\sum_{l\in\delta(d)} w_{l}w_{l-d} x_{t-l+d}^2}_{\Dcal_3(\bxbar)}
\end{align*}

\begin{align*}
  \Dcal_1(\bxbar) &= \sum_{l\in \Lcalbar} \sum_{m\le t \le T} w_l^2 x_{t-l}^2 
            = \sum_{l\in \Lcalbar} \sum_{m-l\le t \le T-l} w_l^2 x_{t}^2 
            = \sum_{1\le t\le T} \rbr{\sum_{l\in\Lcalbar} w_l^2 [m\le t+l\le T]}x_{t}^2 \\
            &= \sum_{1\le t \le T} \rbr{\sum_{l,l'\in\Lcalbar} w_l w_{l'} 
            [m\le t+l\le T][l' = l]} x_t^2 \\
  \Dcal_2(\bxbar) &= \sum_{m\le t \le T}\sum_{1\le d\le L}\sum_{l\in\delta(d)} w_{l}w_{l-d} x_{t-l}^2
  = \sum_{1\le t \le T} \rbr{\sum_{1\le d\le L} \sum_{l\in\delta(d)} w_l w_{l-d}[m\le t + l \le T]} x_t^2 \\
  &= \sum_{1\le t\le T} \rbr{\sum_{\substack{l,l'\in \Lcalbar}} w_l 
  w_{l'}[m\le t + l \le T][l' < l]} x_t^2\\
  \Dcal_3(\bxbar) &= \sum_{m\le t \le T}\sum_{1\le d\le L}\sum_{l\in\delta(d)} w_{l}w_{l-d} x_{t-l+d}^2
  = \sum_{1\le t \le T} \rbr{\sum_{1\le d\le L} \sum_{l\in\delta(d)} w_l w_{l-d}[m\le t + l - d\le T]} x_t^2 \\
  &= \sum_{1\le t\le T} \rbr{\sum_{\substack{l',l\in \Lcalbar}} w_l 
  w_{l'}[m\le t + l \le T][l' > l]} x_t^2
\end{align*}
Let $D \in R^{T\times T}$ be a diagonal matrix with $D_{tt}$ be
the coefficient associated with $x_{t}^2$ in $\Dcal(\bxbar)$.   
Combining the results of $\Dcal_1(\bxbar), \Dcal_2(\bxbar)$, and $\Dcal_3(\bxbar)$, $D_t$ can 
be written as follows.  
\begin{align*}
  D_{tt} = \rbr{\sum_{l\in\Lcalbar} w_l}\rbr{\sum_{\substack{l\in \Lcalbar}} 
  w_l[m\le t+l \le T]}\quad \forall t.
\end{align*}
It is clear that $\Dcal(\bxbar) =\bxbar^\top D \bxbar$. Note that 
$\forall t = m,\ldots,T-L$, $D_{tt} = \rbr{\sum_{l\in\Lcalbar} w_l}^2$.  
\end{proof}
\subsection{Proof of Corollary~\ref{thm:ar-invcov}}
\label{sec:proof-ar-invcov}
\begin{proof}
  It is well known that graph regularization can be written in the
  quadratic form~\cite{AS03a} as follows.
  \[
    \frac{1}{2} \sum_{t\sim s} G_{ts} \rbr{x_t - x_s}^2 = \bxbar^\top \Lap{G} \bxbar,
  \]
  where $\Lap{G}$ is the $T\times T$ graph Laplacian for $G$ defined as:
  \[
    \Lap{G}_{ts} = \begin{cases}
      \sum_{j} G_{tj}, &t = s \\
      -G_{ts}, & t\neq s \text{ and there is an edge } t \sim s \\
      0, &\text{otherwise.}
    \end{cases}
  \]
  Based on the above fact and the results from Theorem~\ref{thm:ar-graph}, we 
  obtain the quadratic form for $\ARr{\bxbar}$ as follows.
  \[
    \ARr{\bxbar} = \frac{1}{2}\bxbar^\top \rbr{\Lap{\GAR} + \underbrace{ D + 
    \eta I}_{\text{diagonal}}} \bxbar.
  \]
  Because $D+\eta I$ is diagonal, the non-zero pattern of the off-diagonal 
  entries of the inverse covariance $\Sigma^{-1}$ for $\ARr{\bxbar}$ is 
  determined by $\Lap{\GAR}$ which shares the same non-zero pattern as $\GAR$.
\end{proof}
\section{Details: Scalability Issue of R-DLM package}
\label{sec:rdlm}
In this section, we show the source code demonstrating that R-DLM fails to 
handle high-dimensional time series even with $n=32$. Interested readers can 
run the following R code to see that the \verb!dlmMLE()! function in R-DLM is 
able to run on a $16$-dimensional time series. However, when we increase the 
dimension to $32$, \verb!dlmMLE()! crashes the entire R program.

\begin{verbatim}
library(dlm)
builderFactory <- function(n,k) {
    n = n;
    k = k;
    init = c(rep(0,k), rep(0.1,3),0.1*rnorm(n*k), 0.1*rnorm(k*k))
    build = function(x) {
        m0 = x[1:k]
        C0 = (abs(x[k+1]))*diag(k)
        V  = (abs(x[k+2]))*diag(n)
        W  = (abs(x[k+3]))*diag(k)
        FF = matrix(nrow=n,ncol=k, data=x[(k+3+1):(k+3+n*k)])
        GG = matrix(nrow=k,ncol=k, data=x[(k+3+n*k+1):(k+3+n*k+k*k)])
        return (dlm( m0=m0, C0=C0, FF=FF, GG=GG, V=V, W=W))
    }
    return (list(n=n,k=k,init=init,build=build))
}

Rdlm_train <- function(Y, k, maxit) {
    if(missing(maxit)) { maxit=10 }

    if(ncol(Y)==3) {
        Ymat = matrix(nrow=max(Y(,1)),ncol=max(Y(,2)))
        Ymat[cbind(Y(,1),Y(,2))] = Y(,3)
    } else {
        Ymat = Y;
    }
    n = nrow(Ymat)
    TT = ncol(Ymat)
    dlm_builder = builderFactory(n, k)
    mle = dlmMLE(Ymat,dlm_builder$init,build=dlm_builder$build,control=list(maxit=10))
    dlm = dlm_builder$build(mle$par)
    dlm_filt = dlmFilter(Ymat,dlm)
    return (dlm_filt)
}

tmp = t(as.matrix(Nile)); 
tmp=rbind(tmp,tmp); tmp=rbind(tmp,tmp); 
tmp=rbind(tmp,tmp); tmp=rbind(tmp,tmp); 

print(nrow(tmp))
Rdlm_train(tmp,4);
print('works')

tmp=rbind(tmp,tmp); 
print(nrow(tmp))
Rdlm_train(tmp,4);

\end{verbatim}

\end{document}